\documentclass[11pt]{article}

\usepackage[final]{acl}

\usepackage{times}
\usepackage{latexsym}

\usepackage[T1]{fontenc}

\usepackage[utf8]{inputenc}

\usepackage{CJKutf8}

\usepackage{microtype}

\usepackage{inconsolata}
\usepackage{multirow}

\usepackage{graphicx}

\title{Good/Evil Reputation Judgment of Celebrities\\
by LLMs via Retrieval Augmented Generation\thanks{
This work has been submitted to the IEEE for possible publication.
Copyright may be transferred without notice, after which this version
may no longer be accessible.
}}

\author{Rikuto Tsuchida\ \ \ \ Hibiki Yokoyama \ \ \ \  Takehito Utsuro\\
  {Deg. Prog. Sys.\&Inf. Eng., Grad. Sch. Sci.\&Tech., University of Tsukuba} \\
  \texttt{\{s2110466, s2320808\}\_@\_u.tsukuba.ac.jp}, \  
  \texttt{utsuro\_@\_iit.tsukuba.ac.jp}
}

\begin{document}
\maketitle
\begin{abstract}
The purpose of this paper is to 
examine whether large language models (LLMs) can understand 
what is good and evil
with respect to judging good/evil reputation of celebrities. 
Specifically, we first apply a large language model 
(namely, ChatGPT)
to the task of collecting sentences that mention 
the target celebrity from articles about celebrities on Web pages.
Next, the collected sentences are categorized based on their contents
by ChatGPT, 
where ChatGPT assigns a category name to
each of those categories.
Those assigned category names are referred to as 
``aspects'' of each celebrity. 
Then, by applying the framework of retrieval augmented generation (RAG),
we show that the large language model
is quite effective in the task of judging good/evil 
reputation of aspects and descriptions of each celebrity. 
Finally,
also in terms of proving the advantages of the proposed method
over existing services incorporating RAG functions, 
we show that 
the proposed method of
judging good/evil of aspects/descriptions of each celebrity 
significantly outperform an existing service incorporating RAG functions.
\end{abstract}

\section{Introduction}
\label{sec:intro}

This paper proposes a method of
judging good/evil 
reputation of celebrities based on information collected from Web pages
by employing large language models (LLMs, 
in particular ChatGPT) via retrieval augmented generation
(RAG)~\cite{rag}.
Efforts have been made to ensure that ChatGPT does not output insults about people, 
but it remains unclear 
whether it can distinguish between good and evil 
or understand the degree of evilness in its content. 
Thus, this paper shows that ChatGPT can judge good/evil reputation 
of celebrities based on their aspects and descriptions.
We also show that
ChatGPT can further distinguish the degree of evilness 
such as illegal or legal but unethical.

However, gathering information about celebrities using 
ChatGPT alone does not necessarily yield the latest 
information. This is because ChatGPT cannot grasp events
that occurred after its training data cut-off. 
Therefore, this paper proposes a method where information 
obtained from external sources is provided to a large language model such as
ChatGPT, allowing it to 
consider events that happened after its training data 
cut-off when generating responses. This approach 
is known as retrieval augmented generation (RAG)~\cite{rag}. 
Also in previous research~\cite{yokoyama24}, impressions regarding 
celebrities were collected and aggregated from posts on platform X. 
The aspects refers to what the impressions are about.
In contrast, this paper proposes a novel method of
extracting aspects of celebrities from Web pages and aggregating 
detailed descriptions of those aspects.
Through the experimental evaluation,
in terms of the variety of aggregated aspects/descriptions, 
we showed that this novel approach is especially effective 
in the case of celebrities who have encountered 
a certain kind of scandals. 

Finally, a comparison is made with Microsoft Copilot\footnote{
    \url{https://www.microsoft.com/ja-jp/microsoft-copilot/organizations}
    }
developed by Microsoft Corporation.
Microsoft Copilot combines ChatGPT with RAG technology,
while its comparison
with the method proposed in this paper
reveals the following:
i.e., 
the proposed method outperforms Microsoft Copilot 
both in the number of aggregated aspects/descriptions 
of celebrities as well as their accuracy. 
This is mainly because 
the proposed method collects 
much larger number of Web pages before aggregation 
and then identifies much larger number of 
aspects/descriptions compared with Microsoft Copilot.

The followings give the contribution of this pa
per:
\begin{itemize}
\setlength{\itemsep}{-0.3em}
\item We proposed a method for widely collecting information on celebrities.
\item With the help of RAG function, we showed that ChatGPT can distinguish
      between good and evil and understand the degree of evilness
      of celebrities based on their aspects and descriptions.
\item We showed the advantages of the proposed method compared to existing services,
       Microsoft Copilot by comparing results of good/evil reputation judgment.
\end{itemize}

\section{Related Work}
Research on information aggregation of celebrities 
includes \citet{yokoyama24} as mentioned in the previous section,
as well as research on determining the relationship between 
celebrities and impressions in Microblogs~\cite{nozaki-paclic}, 
and the work on extracting impressions about celebrities' aspects from 
Microblogs~\cite{sugawara}. 
However, these studies do not address celebrities who have been involved in past controversies. 
This paper resolves the issues that arise when such celebrities are included as subjects.

In this paper, we utilize retrieval augmented generation (RAG)~\cite{rag},
which helps to reduce hallucinations in large language models (LLMs) and
stabilize output by referencing externally obtained information.
Related research on RAG includes retrieval-augmented language models~\cite{rag_2} and improvements in reliability, adaptability, and attribution in retrieval-augmented language models~\cite{rag_3}. Additionally, research related to ChatGPT encompasses entity linking~\cite{entity-matching}, dialogue analysis~\cite{taiwa-bunnseki}, and extractive summarization~\cite{youyaku1}.

Furthermore, an important feature of this paper is the judgment 
between good or evil of celebrities' careers using LLMs.
Research involving LLMs and the legal field includes studies verifying the effectiveness of LLMs in the legal field~\cite{LLMandlaw}.

\section{
Aggregating Impressions on Celebrities and their Reasons
from Microblog Posts and Web Search Pages~\cite{yokoyama24}
}
\label{sec:senkoukenkyu}

This paper builds on prior research outlined 
in Section~\ref{sec:intro}, specifically the evaluation of
RAG
in aggregating reasons for posts expressing impressions about celebrities~\cite{yokoyama24}.
In that study, posts containing impressions on
celebrities were collected from X using the large language model ChatGPT. 
From these collected posts,
impressions regarding what aspects of celebrities are extracted as the celebrity's aspect. 
The reasons for the ``celebrity's aspect + impression'' extracted at this stage 
are collected and aggregated using RAG~\cite{rag}. 
Web pages are searched using
``{\it celebrity's aspect + impression}''
as queries, and the resulting Web pages are provided to ChatGPT to identify reasons for the impressions from them. 
The reason for using RAG~\cite{rag} in this process is 
that the information obtained solely from posts rarely 
contains a detailed explanation of the reasons, even if impressions are present.
Therefore, it is necessary to obtain information about the reputation of celebrities 
from external Web pages. 
The details are shown
in Figure~\ref{fig:senkoukenkyu} in the Appendix~\ref{subsec:appendix_1}.

However, in the previous study~\cite{yokoyama24}, 
only celebrities who had not caused any scandal in the past were considered.
Therefore, when data was collected and aggregated on a wider variety of celebrities, 
it was found that posts related to celebrities who had encountered scandals
in the past often lacked a clear aspects, making extraction difficult. This situation is illustrated in Figure~\ref{fig:senkou_mondaiten}
in the Appendix~\ref{subsec:appendix_1}.
In other words, the approach of the previous study is only effective 
for individuals with no past troubles nor scandals, 
but it is considered ineffective for celebrities who have negative 
impressions such as troubles and scandals in their past.

Addressing these issues, this paper aims to study celebrities
who have encountered scandals in the past and to extract the aspects that are not clearly depicted in posts.

\section{
Proposed Approach:
Skipping Microblog Posts but
Aggregating Celebrities' Aspects/Descriptions 
directly collected from Web Pages
}
\label{sec:wadaisyuyaku}

In order to solve the problems caused in the previous study
mentioned in the previous section, 
this paper proposes a method that retrieves all aspects related to celebrities from Web pages, 
without extracting them from microblog posts. 

\begin{figure*}[tbp]
\centering
\includegraphics[width=0.8\linewidth]{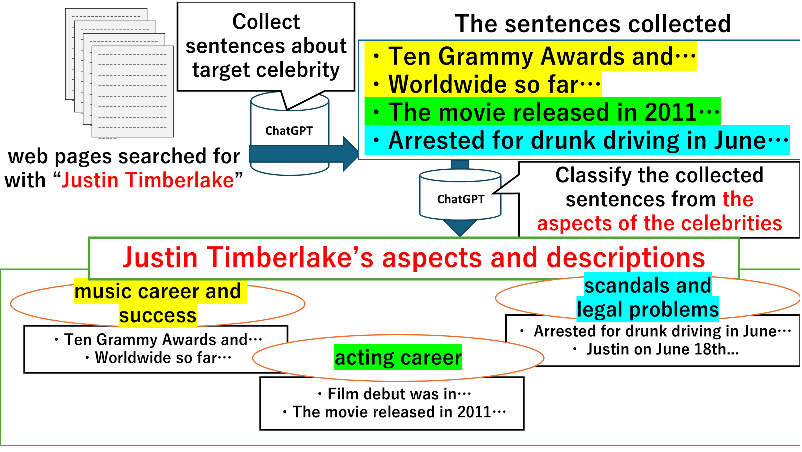}
\caption{An Example of Extracting and Aggregating Aspects and Descriptions for ``Justin Timberlake''}
\label{fig:kanten}
\end{figure*}

\subsection{Collecting Web Pages}
\label{subsec:taisyousya}

First, 20 Web pages are collected by searching with ``
{\it celebrity name}''.
In the previous study~\cite{yokoyama24}, the query for the search was
``{\it celebrity's aspects + impression}'', 
so only Web pages related to the information collected in the microblog posts could be retrieved.
However, 
by revising the query as above, a wide range of information can be retrieved.

Table~\ref{table:data_utiwake} shows the list of celebrities targeted in this paper.
There are ten Japanese celebrities, five of whom were originally included 
in the previous study~\cite{yokoyama24} and have no previous problems. 
The remaining five are new celebrities for study in this paper, 
who have had problems in the past and have been inactive. 
In addition, five additional non-Japanese celebrities who had caused problems were also targeted.
For all of those celebrities for study in this paper, 
we examined Japanese Web pages only, but not the Web pages of
any other language including English Web pages\footnote{
  When collecting Web pages with each of the five non-Japanese 
  celebrities' names as the query, 
  we use the Japanese katakana characters 
  used for those foreign names,
  i.e., 
  ``\begin{CJK}{UTF8}{ipxm}
  ショーン・コムズ
  \end{CJK}''
  (Shoon Komuzu)
  for ``Sean Combs'',
  ``\begin{CJK}{UTF8}{ipxm}
  ケヴィン・スペイシー
  \end{CJK}''
  (Kevin Speisii)
  for ``Kevin Spacey'',
    ``\begin{CJK}{UTF8}{ipxm}
  ジョニー・デップ
  \end{CJK}''
  (Jonii Deppu)
  for ``Johnny Depp'',
  ``\begin{CJK}{UTF8}{ipxm}
  ウィノナ・ライダー
  \end{CJK}''
  (Uinona Raidaa)
  for ``Winona Ryder'', and
    ``\begin{CJK}{UTF8}{ipxm}
  ジャスティン・ティンバーレイク
  \end{CJK}''
  (Jasutein Teinbaareiku)
  for ``Justin Timberlake''.
}. 

\begin{table}
\centering
\scalebox{0.85}{
\begin{tabular}{|c|c|}
\hline
\citet{yokoyama24}
&
only in this paper 
\\
\hline\hline
Ryosuke Yamada & Huwa-chan\\
Kazunari Ninomiya & Pierre Taki \\
Huma Kikuchi & Yuichi Nakamura\\
Syun Oguri& Noriyuki Makihara \\
Go Ayano & Hiroyuki Miyasako\\
\hline
---&Sean Combs\\
---&Kevin Spacey\\
---&Johnny Depp\\
---&Winona Ryder\\
---&Justin Timberlake\\
\hline
\end{tabular}
}
\caption{Celebrities for Study in This Paper}
\label{table:data_utiwake}
\end{table}

\subsection{Extracting and Aggregating Celebrities' Aspects and their Descriptions}
\label{subsec:tyusyutsu}

Next, from the large number of sentences on the collected Web pages, 
sentence that mention the target celebrity are collected, 
and since some of the collected sentence have overlapping contents, 
so they are categorized according to the contents. In this process, 
we ask ChatGPT to determine
``what the topic is''
about the celebrity and name it as a category name. 
The category name generated here becomes
``celebrity's aspects''.
Finally, various descriptions related to the celebrity's aspects are aggregated.

The entire process of collecting sentences that mention celebrities,
categorizing them and creating category names,
and extracting the celebrity's aspects is done using ChatGPT. The model of ChatGPT used is gpt-4o, 
and the same model is used in all the later situations where ChatGPT is used. 
Figure~\ref{fig:kanten} shows the process from collection of Web pages to extraction of aspects 
for ``Justin Timberlake''
as an example of a target celebrity.

\begin{table}[t]
\centering
\tabcolsep 8pt
\scalebox{0.85}{
\begin{tabular}{|c|c|c|}
\hline
\begin{tabular}{c}
celebrity's\\
name
\end{tabular}
& recall & precision\\
\hline\hline
Huwa-chan & 0.75 (6/8) & 1.00 (6/6) \\
Pierre Taki & 0.53 (8/15) & 1.00 (8/8) \\
Yuichi Nakamura & 0.64 (9/14) & 0.90 (9/10) \\
Hiroyuki Miyasako & 0.60 (6/10) & 1.00 (6/6)\\
Noriyuki Makihara & 0.82 (9/11) & 1.00 (9/9)\\
Ryosuke Yamada & 0.78 (7/9) & 1.00 (7/7) \\
Syun Oguri & 0.50 (5/10) & 1.00 (5/5)\\
Go Ayano & 0.71 (10/14) & 0.83 (10/12)\\
Kazunari Ninomiya & 0.63 (5/8) & 0.83 (5/6)\\
Huma Kikuchi & 0.86 (6/7) & 1.00 (6/6) \\
Sean Combs&0.86 (6/7)&1.00 (6/6)\\
Kevin Spacey&0.63 (5/8)&0.83 (5/6)\\
Johnny Depp&0.60 (6/10)&1.00 (6/6)\\
Winona Ryder&0.83 (5/6)&1.00 (5/5)\\
Justin Timberlake&0.78 (7/9)&0.78 (7/9)\\
\hline\hline
macro average& 0.70 & 0.94\\
\hline
\end{tabular}
}
\caption{Evaluation Results of Extracted and Aggregated Aspects and Descriptions}
\label{tab:kanten_table}
\end{table}

\begin{table*}[t]
\centering
\tabcolsep 3pt
\resizebox{0.95\textwidth}{!}{
\begin{tabular}{|l||l|l|}
\hline
\multicolumn{1}{|c||}{
---
}
&
\multicolumn{1}{|c|}{
\begin{tabular}{c}
Aspects/Impressions/Reasons\\
by \citet{yokoyama24}
\end{tabular}
}
&
\multicolumn{1}{c|}{
\begin{tabular}{c}
Aspects/Descriptions\\
by the Proposed Method
\end{tabular}
}
\\
\hline\hline
\begin{tabular}{c}
Aspects/Descriptions and\\
Aspects/Impressions/Reasons\\
\underline{\textbf{overlapping}}\\
between the two methods 
\end{tabular}&
\begin{tabular}{l}
remarks/criticism, rants/criticism, \\ flaming/criticism, flaming/sympathy,\\ dislike/sympathy, dislike/criticism, \\ disliked/sympathy, disappeared/criticism\\
\end{tabular}
&inappropriate remarks and hiatus\\
\hline
\begin{tabular}{c}
Aspects/descriptions and\\
Aspects/Impressions/Reasons\\
\underline{\textbf{not overlapping}}\\
between the two methods
\end{tabular}
&
\multicolumn{1}{|c|}{
None
}
&
\begin{tabular}{l}
career and activities, \\
language skills and educational background, \\
relationships with friends\\
fashion and influence, \\
media appearances\\
\end{tabular}\\
\hline
\end{tabular}
}
\caption{
Aspects/Impressions/Reasons by \citet{yokoyama24}
and 
Aspects/Descriptions
by the Proposed Method
(by ChatGPT, for ``Huwa-chan'')
}
\label{tab:ex-taiou}
\end{table*}

\begin{table*}[t]
\centering
\tabcolsep 8pt
\scalebox{0.85}{
\begin{tabular}{|c||c|c|}
\hline
---&
\begin{tabular}{c}
Aspects/Impressions/Reasons\\
by \citet{yokoyama24}
\end{tabular}
&
\begin{tabular}{c}
Aspects/Descriptions\\
by the Proposed Method
\end{tabular}
\\
\hline\hline
\begin{tabular}{c}
Aspects/Descriptions and\\
Aspects/Impressions/Reasons\\
\underline{\textbf{overlapping}}\\
between the two methods
\end{tabular}
&
5.67 
&
2.00 
\\
\hline
\begin{tabular}{c}
Aspects/Descriptions and\\
Aspects/Impressions/Reasons\\
that \underline{\textbf{not overlapping}}\\
between the two methods
\end{tabular}
&
2.67 
&
5.83 
\\
\hline\hline
Total
&
8.33 
&
7.83 
\\
\hline
\end{tabular}
}
\caption{
Numbers 
of Aspects/Impressions/Reasons by \citet{yokoyama24}
and 
Aspects/Descriptions
by the Proposed Method
(by ChatGPT, averaged over 10 Celebrities)}
\label{tab:taiou}
\end{table*}

\subsection{Evaluation}


The results obtained by ChatGPT are compared with
manually extracted and aggregated reference aspects and descriptions 
for each celebrity name. 
ChatGPT's results were considered a match if both the aspects and descriptions 
were identical or similar in content to reference\footnote{
   In all the results of the experiment, when the aspect is identical or similar to 
   the reference, it is also the case for the description. 
}. 
%

When the set of the human-handled aspects and descriptions is $R(c)$
and the set of the ChatGPT aspects and descriptions is $S(c)$ 
for a certain celebrity $c$,
we define recall and precision as the following formulas. 
{
\[
\mbox{recall}
= \displaystyle{|R(c) \cap S(c)|}
/ \displaystyle{|R(c)|},
\]
\[
\mbox{precision} =  \displaystyle{|R(c)\cap S(c)|}
/ \displaystyle{|S(c)|}
\]
}
Evaluation results are shown
in Table~\ref{tab:kanten_table}.
Recall is approximately 70 \%, and precision is 90 \% or more, 
indicating that the extraction and aggregation are performed accurately.

\begin{figure}[t]
\centering
\includegraphics[width=1\linewidth]{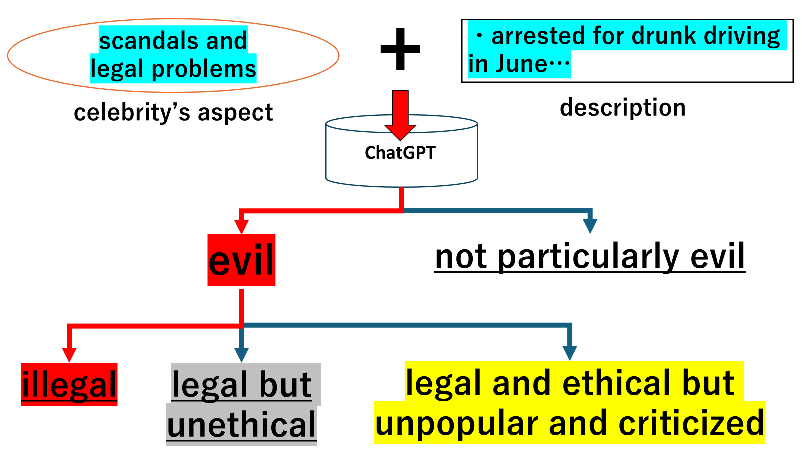}
\caption{
Good/Evil Judgment of a Celebrity's Aspects/Descriptions
(e.g., for ``Justin Timberlake'')}
\label{fig:hantei}
\end{figure}

\begin{table*}
\centering
\tabcolsep 3pt
\scalebox{0.9}{
\begin{tabular}{|c||c|c||c|c|}
\hline
\ \multirow{2}{*}{celebrity's name}
& \multicolumn{2}{c||}{ChatGPT}&\multicolumn{2}{c|}{Microsoft Copilot}\\
\cline{2-5}
&zero-shot&few-shot&zero-shot&few-shot\\
\hline\hline
Huwa-chan &1.00 (6/6)& 1.00 (6/6)&0.70 (7/10)&0.80 (8/10)\\
Pierre Taki &0.86 (7/8)& 1.00 (8/8)&0.60 (3/5)&0.60 (3/5)\\
Yuichi Nakamura &0.78 (7/9)& 1.00 (9/9)&0.83 (5/6)&0.83 (5/6)\\
Hiroyuki Miyasako &0.83 (5/6)& 0.67 (4/6)&0.80(4/5)&0.80 (4/5)\\
Noriyuki Makihara &1.00 (9/9)& 1.00 (9/9)&0.50 (2/4)&0.50 (2/4)\\
Ryosuke Yamada &1.00 (7/7)& 1.00 (7/7)&---&---\\
Syun Oguri & 1.00 (5/5)& 1.00 (5/5)&---&---\\
Go Ayano & 1.00 (10/10)& 1.00 (10/10)&---&---\\
Kazunari Ninomiya &1.00 (5/5)& 1.00 (5/5)&---&---\\
Huma Kikuchi & 1.00 (6/6)& 1.00 (6/6)&---&---\\
\hline
macro average&0.95&0.97&0.67&0.71\\
\hline
\end{tabular}
}
\caption{
Evaluation Results of
Good/Evil Judgment of a Celebrity's Aspects/Descriptions}
\label{tab:hantei_each}
\end{table*}

\begin{table*}[t]
\centering
\tabcolsep 4pt
\scalebox{0.8}{
\begin{tabular}{|c|c||c|c|c|c||c|}
\hline
\multicolumn{2}{|c||}{---}&\multicolumn{5}{c|}{predicted by ChatGPT}\\
\cline{3-7}
\multicolumn{2}{|c||}{---}& illegal & legal but unethical  &
\begin{tabular}{c}
legal and ethical\\
but unpopular and\\
criticized\\
\end{tabular}
& not particularly evil & total \\
 \hline\hline
\multirow{8}{*}{reference}&  illegal &
\begin{tabular}{c}
2\\
Makihara/Taki\\
\end{tabular}
& 0& 0& 0&2\\
 \cline{2-7}
 \cline{2-7}
 &
 \begin{tabular}{c}
 legal\\
 but unethical\\
 \end{tabular}
 &
 \begin{tabular}{c}
 1\\
 Miyasako\\ 
 \end{tabular}
 &
  \begin{tabular}{c}
  2\\
  Huwa/Nakamaru\\ 
 \end{tabular}& 0&0 &3\\
 \cline{2-7}
 &
 \begin{tabular}{c}
legal and ethical\\
but unpopular and\\
criticized\\
\end{tabular}
 & 0& 0&
  \begin{tabular}{c}
 1\\
 Miyasako\\ 
 \end{tabular}& 0&1\\
 \cline{2-7}
 & not particularly evil& 0& 1& 0& 64&65\\
 \cline{2-7}\cline{2-7}
 & total & 3& 3& 1&64 &71\\
 \hline
\end{tabular}
}
\caption{
Confusion Matrix of 
Good/Evil Judgment of Celebrities' Aspects/Descriptions
(for 10 Celebrities)}
\label{tab:hantei}
\end{table*}

\subsection{
Comparison with \citet{yokoyama24}
}
\label{subsec:taiou}


Some of 
the aspects and descriptions extracted by the proposed method overlap with the 
aspects, impressions, and reasons extracted by
\citet{yokoyama24}.
Here, however, 
there exist
those that are extracted only by the proposed method, 
while conversely, 
others are extracted only by \citet{yokoyama24}. 
Therefore, we manually map the
``aspects + impressions + reasons for impressions'' 
obtained by \citet{yokoyama24} to those ``aspects + descriptions'' 
obtained by the proposed method, and 
examined their overlap. 

Table~\ref{tab:ex-taiou} shows the result obtained 
for ``Huwa-chan''. 
``Huwa-chan'' is a Japanese comedian and
has been criticized for his violent outbursts
against other celebrities on social networking sites and is 
currently on hiatus.
Table~\ref{tab:ex-taiou} shows that there are no aspect, impression, nor reason
that can be obtained only by \citet{yokoyama24}.
On the contrary,
there exist plenty of aspects and descriptions that can be obtained 
only by the proposed method, which means that 
the method proposed in this paper is capable of 
more widely collecting celebrities' aspects and their descriptions. 
The details are provided in Figure~\ref{fig:tougou} of Appendix~\ref{subsec:appendix_2}

The overall result of the mapping 
for 10 celebrities 
is shown in Table~\ref{tab:taiou}. 
On the side of \citet{yokoyama24},
the number of {\it overlapping} 
aspects, impressions, and reasons
is 5.67, which is much larger than 
that of {\it not overlapping} ones.
On the side of the proposed method, on the other hand, 
the magnitude relation of 
{\it overlapping} and {\it not overlapping}
numbers are opposite, 
where 
the {\it not overlapping} number
is much larger than 
the {\it overlapping} number. 
This result again indicates that 
the method proposed in this paper is capable of 
more widely collecting celebrities' aspects and their descriptions. 


\section{Good/Evil Judgment of a Celebrity's Aspects/Descriptions}
\label{sec:ryouhihantei}

Next, a judgment
between good and evil is made for the aspects extracted
in Section~\ref{subsec:tyusyutsu}. 
The purpose of this judgment is to determine 
whether a large-scale language model, 
which has not been trained specifically for legal knowledge, can accurately make fine-grained judgments.

\subsection{The Procedure}
\label{sub:hanteitejun}
Judgment by the ChatGPT is made in two stages
as shown in Figure~\ref{fig:hantei}.
First, the celebrities' aspects and descriptions are classified into two categories: 
``evil'' or ``not particularly evil'', 
and then the aspects and descriptions judged to be 
``evil'' are further classified into three categories
as below:

\begin{itemize}
\setlength{\itemsep}{-0.3em}
\item not particularly evil
\item evil
\begin{itemize}
\item illegal
\item legal but unethical
\item legal and ethical\\ but unpopular and criticized
\end{itemize}
\end{itemize}

In particular, with regard to the three categories of
{\it evil}, 
{\it illegal}
is defined to be
a person who clearly violates a law,
{\it legal but unethical}
is defined to be a person who does
not violate any law but does say or does something 
ethically problematic and is criticized by the public,
and 
{\it legal and ethical but unpopular and criticized}
is defined to be a person who does nothing particularly evil 
but has a poor reputation among others.
We will investigate whether ChatGPT can recognize 
the clear distinction among those definitions.

In addition, before judgment,
ChatGPT is input with the aspects and descriptions of the celebrities 
extracted and aggregated in section~\ref{subsec:tyusyutsu}, 
and the judgment is made by referring to those information.
Figure~\ref{fig:hantei} shows the judgment made on 
``Scandals and legal problems'', 
one of the aspects extracted when ``Justin Timberlake''
was the target.
In Figure~\ref{fig:hantei},
the aspect of 
``Scandals and legal problems'' are
entered into ChatGPT together
with the aggregated description, and ChatGPT judges where it fits in the classification 
by referring to the aggregated description. 
In the first stage, 
``Scandals and legal problems'' was judged to be
``evil'' at the first stage, and as the result of the second stage of 
the judgment,
``Scandals and legal problems'' of 
``Justin Timberlake'' was judged to be ``illegal''.

These decisions were made by zero-shot and few-shot.
The examples used in the prompt are not related to the actual celebrities for study.
One example was created for each of the four categories, for a total of four examples
and are used as the four-shot. 


\subsection{Evaluation Results}

For each of the extracted and aggregated aspects and descriptions
from an celebrity, we examined 
whether the results of the ChatGPT matched the results of manual reference
for good/evil judgments. 
Table~\ref{tab:hantei_each} shows the evaluation results 
for each of the 10 celebrities for study. 
In addition, a confusion matrix summarizing the results for the 
10 celebrities is shown in Table~\ref{tab:hantei}. 
In Table~\ref{tab:hantei}, the names of the celebrities
whose aspects/descriptions are 
classified into one of the three ``evil'' categories are shown. 
Out of the 10 celebrities, 
all of 
the five shown in Table~\ref{tab:hantei}
have been suspended from the entertainment industry due to
scandals.
Thus, it can be said that accurate judgments have been made\footnote{
   As the few-shot, four more examples were added,
    for a total of eight examples as few-shot, 
   where the results did not change.
    Therefore, the results in all tables are on four examples 
    as few-shot.
}.

\begin{table*}
\centering
\scalebox{0.84}{
\begin{tabular}{|c||l|l||l|}
\hline
\begin{tabular}{c}
celebrity's name\\
(The year the scandal occurred,\\
or ``---'' 
in the case of no scandal
)\\
\end{tabular}
&
\multicolumn{1}{c|}{
without RAG
}
&
\multicolumn{1}{c||}{
with RAG
}
&
\multicolumn{1}{c|}{
reference
}
\\
\hline\hline
Yamaguchi Tatsuya (2018/2020)&T / illegal&T / illegal&illegal\\
Pierre Taki (2019)&T / illegal&T / illegal&illegal\\
Hiroyuki Miyasako (2019)& T / legal but unethical &T / legal but unethical &legal but unethical \\
Noriyuki Makihara (2020)&T / illegal&T / illegal&illegal\\
Johnny Kitagawa (September 2023)&T / illegal&T / illegal&illegal\\
\hline
Hitoshi Matsumoto (December 2023)&
\underline{\bf F}
/ not particularly evil&T / legal but unethical &legal but unethical \\
Sinji Saito (December 2023)&
\underline{\bf F}
/ not particularly evil&T / illegal&illegal\\
Toshihumi Hujimoto (2024)&
\underline{\bf F}
/ not particularly evil&T / illegal&illegal\\
Huwa-chan (2024)&
\underline{\bf F}
/ not particularly evil&T / legal but unethical &legal but unethical \\
Yuichi Nakamura (2024)&
\underline{\bf F}
/ not particularly evil&T / legal but unethical &legal but unethical \\
Masahiro Nakai (2024)&
\underline{\bf F}
/ not particularly evil&
\underline{\bf F}
/ legal but unethical &illegal\\
\hline
Ryosuke Yamada (---)&T / not particularly evil&T / not particularly evil&not particularly evil\\
Kazunari Ninomiya (---)&T / not particularly evil &T / not particularly evil&not particularly evil \\
Go Ayano (---)&T / not particularly evil &T / not particularly evil &not particularly evil \\
Huma Kikuchi (---)&T / not particularly evil &T / not particularly evil &not particularly evil \\
Syun Oguri (---)&T / not particularly evil &T / not particularly evil &not particularly evil \\
Kuro-chan (---)&
\multicolumn{1}{p{3cm}|}{
T / legal and ethical
but unpopular and
criticized
}
&
\multicolumn{1}{p{3cm}||}{
T / legal and ethical
but unpopular and
criticized
}
&
\multicolumn{1}{p{2.7cm}|}{
legal and ethical
but unpopular and
criticized
}
\\
\hline\hline
Winona Ryder (2001)&T / illegal&T / illegal&illegal\\
Johnny Depp (2016)&T / not particularly evil &T / not particularly evil &not particularly evil \\
Kevin Spacey (2017)&T / illegal&T / illegal&illegal\\
\hline
Sean Combs (2024)&
\underline{\bf F}
/ not particularly evil&T / illegal&illegal\\
Justin Timberlake (2024)&
\underline{\bf F}
/ not particularly evil&T / illegal&illegal\\
\hline\hline
total accuracy &
\multicolumn{1}{c|}{
0.64 (14/22)
}
&
\multicolumn{1}{c||}{
0.95 (21/22)
}
&
\multicolumn{1}{c|}{
---
}
\\
\hline
\end{tabular}
}
\caption{
Comparison of with/without RAG
in 
Good/Evil Judgment per Celebrity
by ChatGPT
}
\label{tab:rag_effect}
\end{table*}

\section{
Good/Evil Judgment per Celebrity
(not per aspect/description)}
\label{sec:huhyou}

In the previous section, 
we extract and aggregate aspects and descriptions of the celebrities and 
let ChatGPT judge the distinction of good and evil of each aspect. 
Instead of making this judgment for each aspect,
this section performs 
good/evil judgment per celebrity, 
but not per aspect/description.
Furthermore, 
we compare the judgment results of ChatGPT without RAG 
with those of ChatGPT with RAG, and show that the use of RAG 
improves the accuracy of judgment. 

\subsection{The Procedure}

The information given is each celebrity's aspects and descriptions 
as described in Section~\ref{subsec:tyusyutsu}.
Each celebrity's aspects and descriptions 
are given to ChatGPT as prior information as the celebrity's reputation.
Again, the model used in ChatGPT is gpt-4o.

The detailed definitions of the classifications used for 
good/evil judgment 
are given in Section~\ref{sec:ryouhihantei}.
Furthermore, in order to clarify that the 
date/year of the scandal have influence on the judgment result, 
several celebrities were added to the five 
celebrities studied in the previous section.  
The same survey was also conducted not only on Japanese 
but also on foreign celebrities.

\subsection{Evaluation Results}


The results are shown in Table~\ref{tab:rag_effect}. In Table~\ref{tab:rag_effect}, 
if the correct judgment was made in comparison with reference, 
a ``T'' was entered, and 
otherwise, 
a ``F'' was entered. 
In Table~\ref{tab:rag_effect}, in the case of 
``Justin Timberlake'', the result of judgment without using RAG is 
``not particularly evil'', which is different from the correct label given by human hand. 
Here, the result of having ChatGPT make a judgment after providing it with the reputation on 
``Justin Timberlake'' collected and aggregated from Web pages,
the classification result changed from
``not particularly evil'' to ``illegal'', 
and the incorrect classification was revised to the correct classification by using RAG.

 The results of other celebrities 
 show that, 
 for several celebrities, accurate judgments were made
 even without RAG,
 the results are also the same with RAG. 
 It is interesting to note here
 that those celebrities with correct judgments 
 even without RAG have encountered their scandals 
 in the year of 2023 or before. 
 %
 In other words, the ChatGPT can make accurate judgments about scandals that occurred 
 within the time period of the data used to train ChatGPT, because the information is contained in it. 
 On the other hand, ChatGPT has no knowledge of information after the period of training data,
 so it makes incorrect judgments without any appropriate training information.
 Since the model used in this study, gpt-4o, was trained on data up to October 2023, 
 as shown in Table~\ref{tab:rag_effect}, it can be seen that the classification 
 without RAG is incorrect for celebrities who had scandals after October 2023.

From the above, it can be said that the use of RAG has 
benefits on improving
the accuracy of the judgment.
In other words, it can be said that when collecting and aggregating 
the reputations of celebrities, 
the use of RAG makes it possible to better guarantee their accuracy
and to obtain the correct output when judging distinction between good and evil.

\section{Comparison with Existing Services}
\label{sec:copilot}
Finally, extracted and aggregated aspects and descriptions 
in the proposed method are compared with the results of 
existing services with RAG function
to demonstrate the advantages of the proposed method. 
Microsoft Copilot is used for the comparison. Microsoft Copilot is a combination of the search engines Bing\footnote{\url{https://www.bing.com/?cc=jp}} and gpt-4o, which were provided by Microsoft Corporation. 
In other words, when gpt-4o is used alone,
the output concerning the latest information is not stable and incorrect answers are given, as described 
in Section~\ref{sec:senkoukenkyu} and 
Section~\ref{sec:huhyou}.
However, 
this situation can be avoided through RAG. 
Microsoft Copilot creates search queries from the input text, and answers the questions 
by referring to the information on the retrieved Web pages.
Since Microsoft Copilot uses the RAG~\cite{rag} technology, 
only the initial input is done by a human, 
and the search queries and the selection of Web pages to be retrieved are all done automatically.

\subsection{The Procedure}

In the good/evil judgment conducted in Section~\ref{sec:ryouhihantei}, 
Web pages searched with the query ``{\it celebrity name}'' were collected, 
and aspects and descriptions related to the query celebrity
were extracted.
In this context, Microsoft Copilot was first tasked with referencing celebrity Web pages
to list topics related to the query celebrity. 
Then, using zero-shot or few-shot,
it was asked to judge on those topics. 
The examples used for few-shot were the same 
as those used in Figure~\ref{fig:prompt}
in the Appendix~\ref{app:pro}. 
Additionally, a summary of the results of zero-shot/few-shot is presented 
in Table~\ref{tab:hantei_each}. Even though the number of examples used 
for few-shot was increased to eight,
no changes were observed in the results, 
so Table~\ref{tab:hantei_each} records the results 
from using four examples. 
Furthermore, the judgment by Microsoft Copilot
was conducted solely on the celebrities listed 
in Table~\ref{table:data_utiwake} who had encountered some form of scandal.

\subsection{Comparison Results}
The percentage of correct answers by 
Microsoft Copilot in Table~\ref{tab:hantei_each} 
shows that the overall percentage of correct answers has dropped significantly.
One reason for this is that Microsoft Copilot retrieves fewer Web pages. 
Although we instructed Microsoft Copilot to retrieve 20 Web pages, as in
the case of the proposed method, 
it actually retrieved only about 2 or 3 pages, which may have resulted in biased information.
In addition, although detailed information on each topic is necessary for an accurate
distinction between good and evil, a few Web pages were not enough to obtain sufficient information. 

On the other hand, it took only a few seconds for Microsoft Copilot to collect information on one celebrity, 
while it took about 10 minutes for our proposed method. 
This is because this paper's method requires reading sentences from a large number of Web pages, 
which is time-consuming, 
while Microsoft Copilot is designed to be accessed from all over the world, 
so it cannot read as many as 20 pages, but only 1 or 2 pages.

In other words, Microsoft Copilot uses RAG technology so that 
gpt-4o can refer to the latest information, 
and it is much faster than the method in this paper. 
However, in terms of depth and breadth of the collected information, 
as can be seen from Table~\ref{tab:hantei_each}, 
the method in this paper is able to collect more detailed information.

\section{Conclusion}

By applying the framework of RAG,
we showed that a large language model (i.e., ChatGPT)
is quite effective in the task of judging good/evil 
reputation of aspects and descriptions of each celebrity. 
Especially compared with the method of \citet{yokoyama24},
in terms of the variety of aggregated aspects/descriptions, 
we showed that our novel approach is effective 
in the case of celebrities who have encountered 
a certain kind of scandals. 
%
%
%
Finally, we compared our method with Microsoft Copilot, 
which provides gpt-4o using RAG. The results showed that
while Microsoft Copilot was superior in terms of faster output by utilizing RAG,
our method excelled in the exploration of detailed information and the breadth of information.



\bibliography{latex/my_en}

\appendix

\section{Appendix}
\label{sec:appendix}
\subsection{Previous study}
\label{subsec:appendix_1}
\begin{figure*}[t]
\centering
\includegraphics[width=0.8\linewidth]{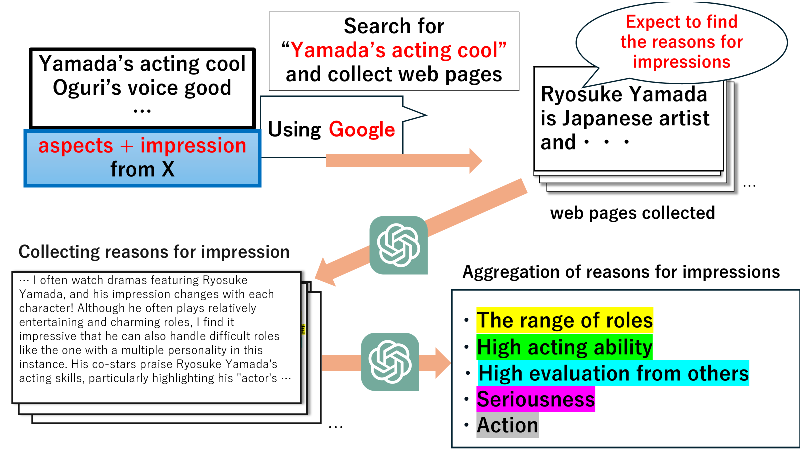}
\caption{Collecting and Aggregating Aspects, Impressions, and Reasons for Impressions 
in the Previous Study~\cite{yokoyama24}}
\label{fig:senkoukenkyu}
\end{figure*}

The method used in previous research~\cite{yokoyama24} is illustrated in Figure~\ref{fig:senkoukenkyu}.

\begin{figure*}[t]
    \centering
    \includegraphics[width=0.7\linewidth]{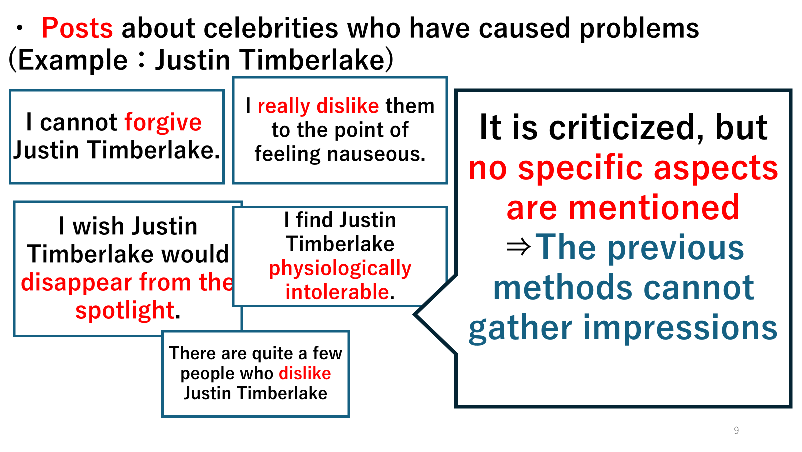}
    \caption{An Example of Extracting and Aggregating Aspects 
    for ``Justin Timberlake'' from Microblog Posts }
    \label{fig:senkou_mondaiten}
\end{figure*}

Figure~\ref{fig:senkou_mondaiten} uses ``Justin Timberlake'' as an example to show that posts
about people who have caused problems in the past do not clearly indicate the aspects. 

\subsection{Set of prompts used}
\label{app:pro}
\begin{figure*}
\centering
\includegraphics[width=1.0\linewidth]{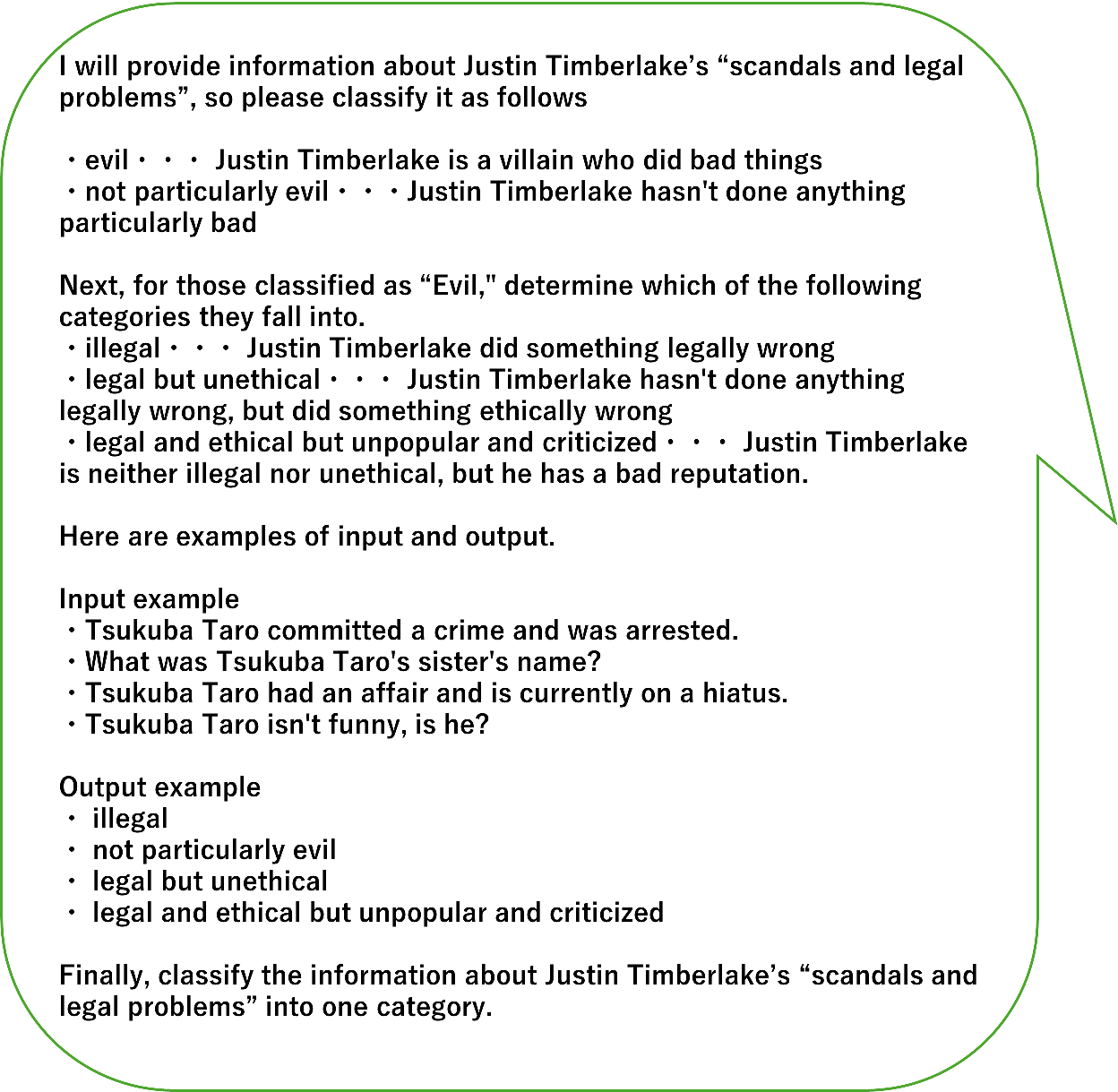}
\caption{The Prompt of ChatGPT for Good/Evil Judgment of a Celebrity's Aspects/Descriptions
(e.g., for ``Justin Timberlake'')
}
\label{fig:prompt}
\end{figure*}

Figure~\ref{fig:prompt} shows the examples and concrete prompts used 
in Section~\ref{sub:hanteitejun}.

\begin{figure*}[t]
\centering
\includegraphics[width=0.9\linewidth]{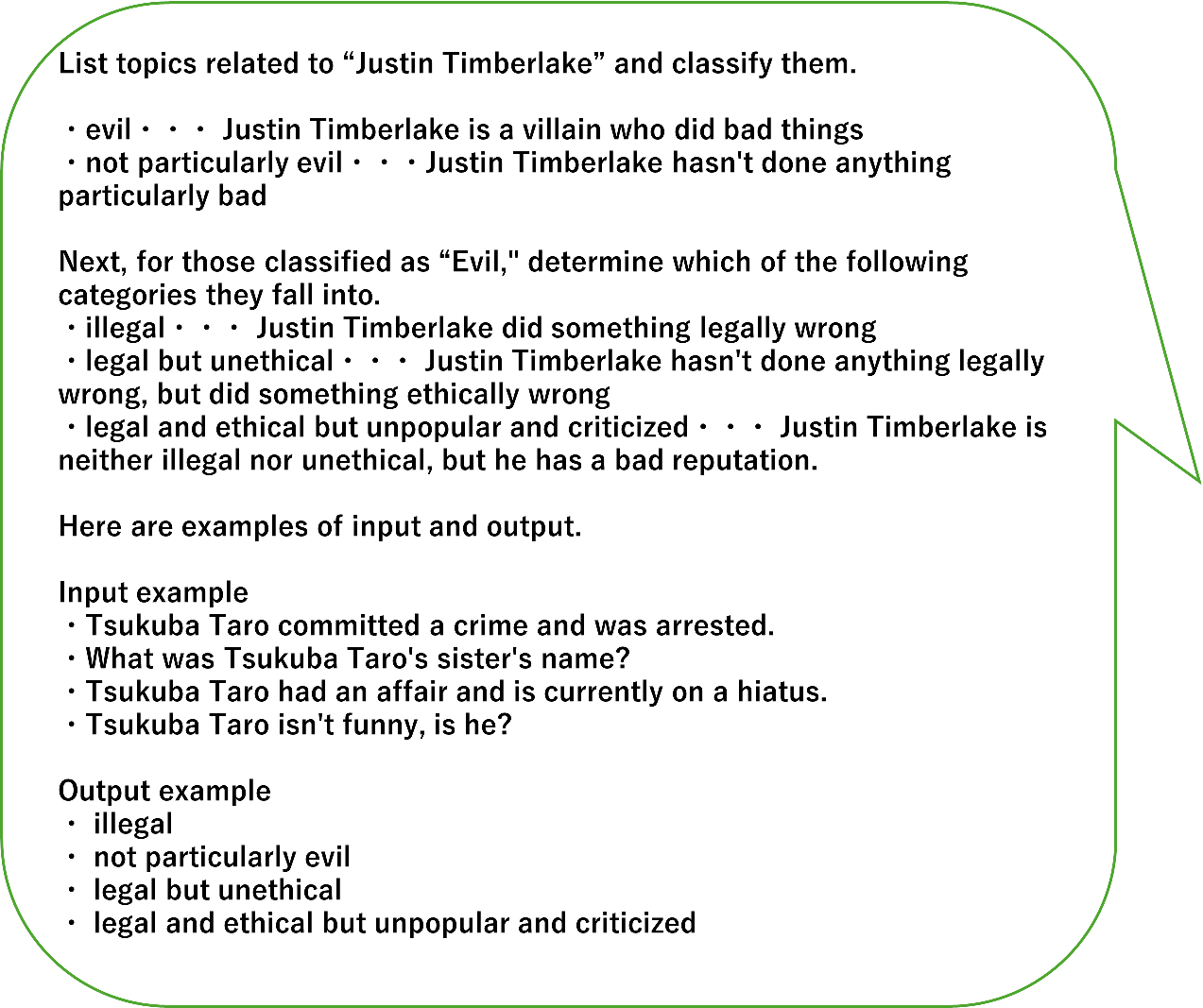}
\caption{The Prompt of Microsoft Copilot for Good/Evil Judgment (e.g., for ``Justin Timberlake'')}
\label{fig:micro_prompt}
\end{figure*}

\begin{figure*}[t]
\centering
\includegraphics[width=1.0\linewidth]{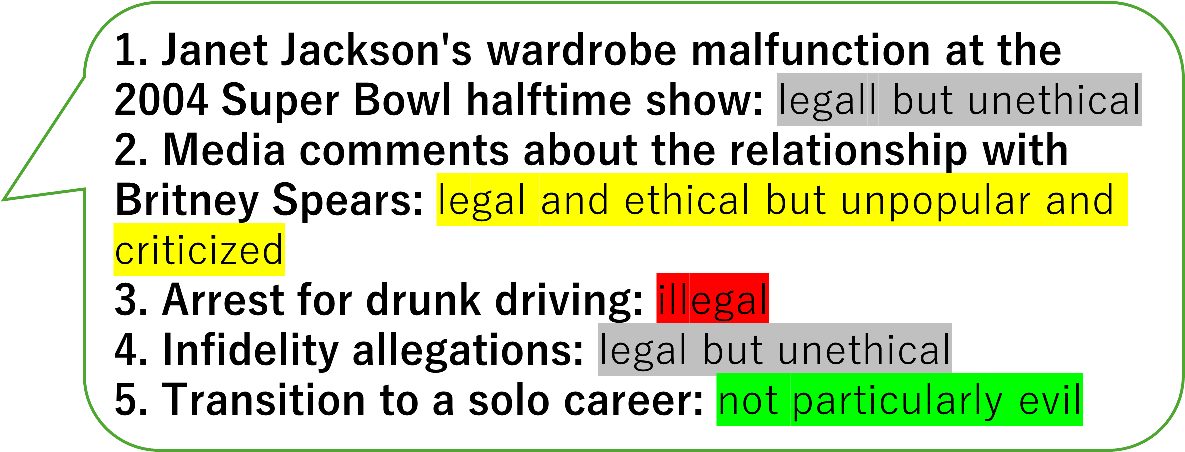}
\caption{An Example of Results by Microsoft Copilot for ``Justin Timberlake''}
\label{fig:micro_r}
\end{figure*}

The exact prompts provided are shown in Figure~\ref{fig:micro_prompt} on judgment
by Microsoft Copilot, and the results of the evaluation 
focusing on the celebrity ``Justin Timberlake'' are shown in Figure~\ref{fig:micro_r}.

\subsection{Correspondence between Aspects/Descriptions obtained by the Method of This Paper
and Aspects/Impressions/Reasons obtained by \citet{yokoyama24}}
\label{subsec:appendix_2}

\begin{figure*}
\centering
\includegraphics[width=1.0\linewidth]{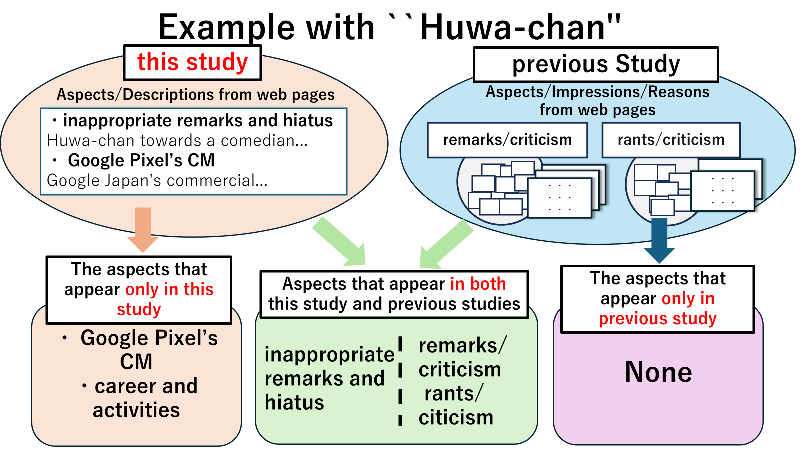}
\caption{An Example of Correspondence between Aspects/Descriptions obtained by the Method of This Paper
and Aspects/Impressions/Reasons obtained by \citet{yokoyama24}
for ``Huwa-chan''}
\label{fig:tougou}
\vspace{9mm}
\end{figure*}

 
 Figure~\ref{fig:tougou}
 shows 
 correspondence between aspects/descriptions obtained by the method of this Paper
and aspects/impressions/reasons obtained by \citet{yokoyama24}.

\subsection{The results of judgment of good or evil by ChatGPT and Microsoft Copilot}
Table~\ref{tab:hantei_detail1} and Table~\ref{tab:hantei_detail2}
show a detailed view of the five celebrities who have encountered a certain 
kind of scandals, 
down to their aspects and the results of their judgments
in Section~\ref{sec:ryouhihantei}.
Moreover, Table~\ref{tab:hantei_detail1} and 
Table~\ref{tab:hantei_detail2} show the aspects of 
each celebrity that 
Microsoft Copilot identified and their judgment results evaluated 
in Section~\ref{sec:copilot}.

\begin{table*}
\centering
\tabcolsep 3pt
\scalebox{0.83}{
\begin{tabular}{|p{18mm}||p{36mm}|l||p{45mm}|p{37mm}|}
\hline
\multirow{2}{*}{
\begin{tabular}{c}
celebrity\\
name
\end{tabular}
}
&\multicolumn{2}{c||}{ChatGPT}&\multicolumn{2}{c|}{Microsoft Copilot}\\
\cline{2-5}
&aspects&results of judgment&aspects&results of judgment\\
\hline\hline
\multirow{19}{*}{Huwa-chan} &inappropriate remarks and hiatus&T / legal but unethical &past problematic behavior&T / legal but unethical \\
&career and activities&T / not particularly evil&freelance activities&T / not particularly evil\\
&language skills and educational background&T / not particularly evil&pro wrestling debut&T / not particularly evil\\
&relationships with friends&T / not particularly evil&statements on social media can cause controversy&T / legal and ethical but unpopular and criticized\\
&fashion and influence&T / not particularly evil&unique fashion&\underline{\bf F} / legal and ethical but unpopular and criticized\\
&&&&($\rightarrow$ not particularly evil)\\
& media appearances&T / not particularly evil&activities on TV shows&T / not particularly evil\\
&&&moving abroad and life there&T / not particularly evil\\
&&&commercial appearances and their impact&T / not particularly evil\\
&&&the creation and success of a YouTube channel&T / not particularly evil\\
&&&dismissed due to trouble with the talent agency&\underline{\bf F} / illegal
($\rightarrow$ legal and ethical but unpopular and criticized)
\\
\hline
\multirow{12}{*}{Pierre Taki} &drug incident and its impact&T / illegal&violation of the narcotics control act&T / illegal\\
&musical activities&T / not particularly evil&musical activities&T / not particularly evil\\
&acting activities and roles in productions&T / not particularly evil&family-oriented&T / not particularly evil\\
&comeback and reception after arrest&T / not particularly evil&media coverage&\underline{\bf F} / legal but unethical ($\rightarrow$ not particularly evil)\\
&impact of television and movies&T / not particularly evil
&impact after arrest&
\underline{\bf F} / legal and ethical but unpopular and criticized\\
&other activities&T / not particularly evil&&
($\rightarrow$ not particularly evil)\\
&hobbies&T / not particularly evil&
&\\
&writing activities&T / not particularly evil&&\\
\hline
\end{tabular}
}
\caption{Results of Good/Evil Judgments on Celebrities' Aspects and Descriptions (1)}
\label{tab:hantei_detail1}
\vspace{5mm}
\end{table*}

\begin{table*}
\centering
\tabcolsep 3pt
\scalebox{0.85}{
\begin{tabular}{|c||p{40mm}|p{36mm}||p{40mm}|p{36mm}|}
\hline
\multirow{2}{*}{
\begin{tabular}{c}
celebrity\\
name
\end{tabular}
}
& 
\multicolumn{2}{c||}{ChatGPT}&\multicolumn{2}{c|}{Microsoft Copilot}\\
\cline{2-5}
&aspects&results of judgment&aspects&results of judgmen\\
\hline\hline
\multirow{17}{*}{
\begin{tabular}{c}
Yuichi\\
Nakamaru\\
\end{tabular}
}
&scandal&T / legal but unethical &graduated from Waseda University&T / not particularly evil\\
&manga artist&T / not particularly evil&human beatbox&T / not particularly evil\\
&YouTube activities&T / not particularly evil&YouTube activities&T / not particularly evil\\
&activities as a member of KAT-TUN&T / not particularly evil&reports of a secret meeting with a female college student&T / legal but unethical \\
&being late&T / legal and ethical but unpopular and criticized&being late&T / legal and ethical but unpopular and criticized\\
&special skills&T / not particularly evil&desire to return after infidelity reports&\underline{\bf F} / legal and ethical but unpopular and criticized\\
&educational background&T / not particularly evil&&($\rightarrow$ not particularly evil)\\
&marriage&T / not particularly evil&&\\
&activities on TV show&T / not particularly evil&&\\
\hline
\multirow{12}{*}{
\begin{tabular}{c}
Hiroyuki\\
Miyasako\\
\end{tabular}
}&YouTube activities&T / not particularly evil&YouTube activities&T / not particularly evil\\
&restaurant management&T / not particularly evil&activities as a businessperson&T / not particularly evil\\
&problem of
underground business dealings
&\underline{\bf F} / illegal
($\rightarrow$ legal but unethical )
&problem of underground business dealings
&\underline{\bf F} / illegal
($\rightarrow$ legal but unethical )\\
&human relationships&T / not particularly evil&allegations of infidelity&T / legal but unethical \\
&music activities&T / not particularly evil&statements on YouTube&T / legal and ethical but unpopular and criticized\\
&TV appearances and comeback&\underline{\bf F} / legal but unethical ($\rightarrow$ not particularly evil)&&\\
\hline
\multirow{14}{*}{
\begin{tabular}{c}
Noriyuki\\
Makihara\\
\end{tabular}
}
&musical activities and hit songs&T / not particularly evil&musical activities&T / not particularly evil\\
&legal problems and arrest record&T / illegal&arrested for violating the stimulant drugs control act&T / illegal\\
&music production techniques&T / not particularly evil&drug use&\underline{\bf F} / legal but unethical ($\rightarrow$ illegal)\\
&evaluation by other artists&T / not particularly evil&arrest record&\underline{\bf F} / legal and ethical but unpopular and criticized\\
&transfer&T / not particularly evil&&($\rightarrow$ illegal)\\
&personal information&T / not particularly evil&&\\
&animal lover&T / not particularly evil&&\\
&resumption of activities&T / not particularly evil&&\\
&album and reviews&T / not particularly evil&&\\
\hline
\end{tabular}
}
\caption{Results of Good/Evil Judgments on Celebrities' Aspects and Descriptions (2)}
\label{tab:hantei_detail2}
\end{table*}

\end{document}